\documentclass[doc]{apa6}

\usepackage{listings}
\usepackage{float}

\usepackage{amsmath,amssymb}
\usepackage{mathtools}

\usepackage{changepage}

\usepackage[utf8x]{inputenc}

\usepackage{textcomp,marvosym}

\usepackage[numbers, super, sort]{natbib}

\usepackage[table]{xcolor}

\usepackage{array}

\usepackage{subcaption}

\title{Reservoir Computing for Fast, Simplified Reinforcement Learning on Memory Tasks}
\shorttitle{Reinforcement Learning with Reservoir Computers}
\author{Kevin McKee}
\affiliation{Astera Institute}

\abstract{
Tasks in which rewards depend upon past information not available in the current observation set can only be solved by agents that are equipped with short-term memory.
Usual choices for memory modules include trainable recurrent hidden layers, often with gated memory.
Reservoir computing presents an alternative, in which a recurrent layer is not trained, but rather has a set of fixed, sparse recurrent weights.
The weights are scaled to produce stable dynamical behavior such that the reservoir state contains a high-dimensional, nonlinear impulse response function of the inputs.
An output decoder network can then be used to map the compressive history represented by the reservoir's state to any outputs, including agent actions or predictions. 
In this study, we find that reservoir computing greatly simplifies and speeds up reinforcement learning on memory tasks by (1) eliminating the need for backpropagation of gradients through time, (2) presenting all recent history simultaneously to the downstream network, and (3) performing many useful and generic nonlinear computations upstream from the trained modules.
In particular, these findings offer significant benefit to metalearning that depends primarily on efficient and highly general memory systems.
}

\begin{document}
\maketitle
\section*{Introduction}
Deep reinforcement learning (RL) algorithms\cite{mnih2015human} typically entail greater training complexity and fragility of solutions than other domains of machine learning.
The difficulty is compounded when the task can only be solved by retaining a medium to long-term memory of the observations, a form of partially observed Markov decision process (POMDP).
If the agent is equipped with a trainable memory module, then reward and error signals must be back-propagated over as many steps of processing is necessary to connect actions with their delayed outcomes.
Gated memory modules such as the Gated Recurrent Unit (GRU)\cite{cho2014learning} and Long Short Term Memory (LSTM)\cite{hochreiter1997long} mitigate the vanishing and exploding gradient problem, but rely on chance initialization of weights to produce an initial memory signal that training can enhance.
As a result, gated memory units can be unreliable and sensitive to hyperparameters and initial conditions when used for RL.

Reservoir computing presents an alternative approach to memory that may reduce the training complexity and improve the reliability of solutions for RL.
A reservoir computer is any high-dimensional, nonlinear dynamical system used to represent, model, and predict a time series of inputs.
The dynamical system typically does not have any free parameters and does not need to be trained.
Inputs are framed as unique perturbations to the system, and training is limited to just a final mapping of the resulting trajectories to the final outputs.
The reservoir serves two purposes: nonlinear expansion of the input perturbations to a high dimensional space, and to retain traces of the input activity over time.
By expanding the dimensionality of the inputs, a trainable feed-forward network, known as the decoder, can recombine those dimensions into useful features.
By retaining traces of activity over time, those features can represent the temporal behavior of the inputs.\cite{lukovsevivcius2009reservoir, tanaka2019recent}

In practice, the output decoder may be nothing more than a linear model that takes the reservoir state as input.
However, this requires that the reservoir performs a sufficient set of nonlinear transformations, which cannot be guaranteed if the reservoir is not extremely large.
An alternative is to use a nonlinear decoder, such as a multi-layer perceptron (MLP), which will directly perform any necessary nonlinear transformations prior to output.

Reservoir computers have been known under two names: Echo State Networks (ESN),\cite{jaeger2004harnessing, jaeger2001short, jaeger2001echo} which use simple recurrent neural networks with the $\tanh$ activation as a basis, and Liquid State Machines (LSM),\cite{maass2002real} which use spiking neurons. 
Because the underlying principle is so general, others have exploited exotic hardware choices to replace the recurrent network, such as a literal reservoir of water.\cite{fernando2003pattern}
For simplicity and computational efficiency, we used Echo State Networks for this study.

In this study, two ESN designs are compared to several standard memory choices for RL agents on diagnostic tasks that characterize a wide variety of memory-based problems, including probability and sequence metalearning, navigation, and tasks requiring arbitrary intervals of recall.
Although the modeling strategy presented here is not new, it has not been widely recognized as a competitive and realistic choice.
For example, Morad et al. tested a suite of RL models across several POMDPs and concluded that the GRU performed best overall, but did not test any form of reservoir computer. \cite{morad2023popgym}
Subramoney et al. demonstrate that reservoirs consisting of spiking neurons can be trained to meta-learn but did not specifically test the performance against common recurrent models.\cite{subramoney2021reservoirs}

\section*{Tasks}
\paragraph{Recall Match}
This task tests the capability of the network to learn outputs that depend exclusively on relations between memories at different times. 
At every step the agent receives one of $K$ symbols.
The agent must output 1 if the symbol at time $t-n$ is the same as the symbol at time $t-m$ for $n\neq m$.
Although this task is simple, it may be difficult for memory modules that require first learning what and when to remember.
In this experiment, $n$ and $m$ were fixed to 2 and 4 and each episode consisted of 100 steps.

\paragraph{Multi-armed bandit}
The multi-armed bandit tests whether probabilistic inferences and associated decisions can be encoded in reservoir working memory and reinstated much later in time.
Each episode consists of random reward for choosing one of $k$ actions.
The probabilities of reward per choice are chosen randomly at the beginning of each episode.
The reward is sampled from a categorical distribution over the reward probabilities (as opposed to independent Bernoulli variables).
The difficulty settings are as follows: 2 bandit arms, 100 steps per episode.

\paragraph{Sequential bandits}
This task tests whether multi-step decision processes can be discovered and exploited via working memory.
In this task, reward is given only for performing a certain sequence of actions.
Each rewarding sequence is randomly chosen by the environment and paired with a random cue.
The rewards per sequence were not probabilistic and only one ordering of actions per episode resulted in reward, meaning the agent could immediately exploit the rewarding sequence after finding it. 
Success on this task represents a primitive form of planning or automation in which the working memory context vector determines several steps of actions into the future.
The difficulty settings are as follows:
2 bandit arms and 3 decision steps making 8 possible sequences. 30 attempts per episode.

\paragraph{Water Maze}
This task demonstrates storage and retrieval of navigational information, sometimes called ``mental maps'' in working memory alone.
It is based on the experimental paradigm introduced by Morris et al., \cite{morris1982place} in which rats are placed in a circular tub of water with symbols posted on the walls.
The rats explore the tub to re-locate a platform hidden just below the surface of the water.
Upon repeated trials, they must use the posted symbols as spatial and directional landmarks to re-locate the platform.
The agent's field of view is the 8 cells of the grid neighboring itself, which were binary coded to indicate an impassable wall.
Hence, in this experiment, the logic of the task is the same, but the unique landmarks are the walls and four corners of the grid.
The agent begins at a random location on a $n\times n$ grid.
The agent gets multiple attempts from different, random start locations to search and find the reward on the grid, which is never the same as the start location.
Once found, the agent can use any remaining trials to move directly to the reward based on navigational instructions retained in working memory.
The difficulty settings are as follows:
4x4 grid, 30 steps per episode, 5 episodes per target.

\section*{Models}
\begin{figure}[!t]
    \centering
    \includegraphics[width=.9\linewidth]{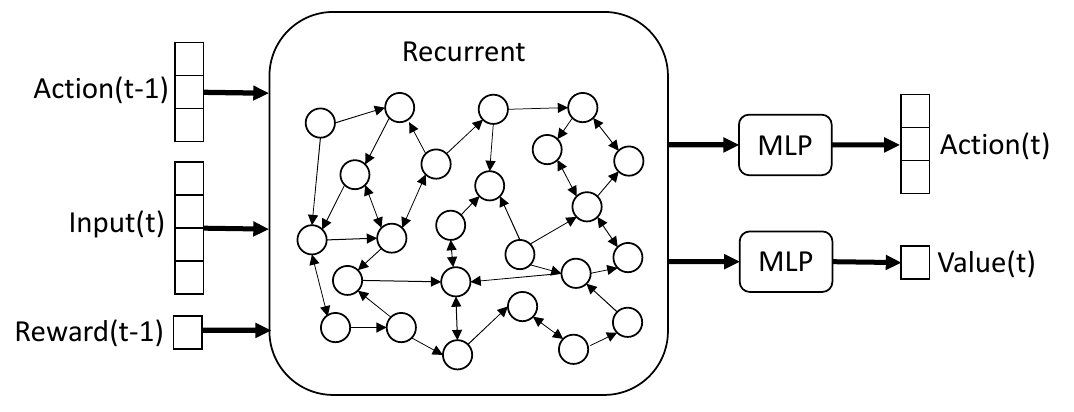}
    \caption{Agent design: inputs and feedbacks are passed into the recurrent module, which then feeds forward to actions and values.}
    \label{fig:modeldiagram}

    \includegraphics[width=0.8\linewidth]{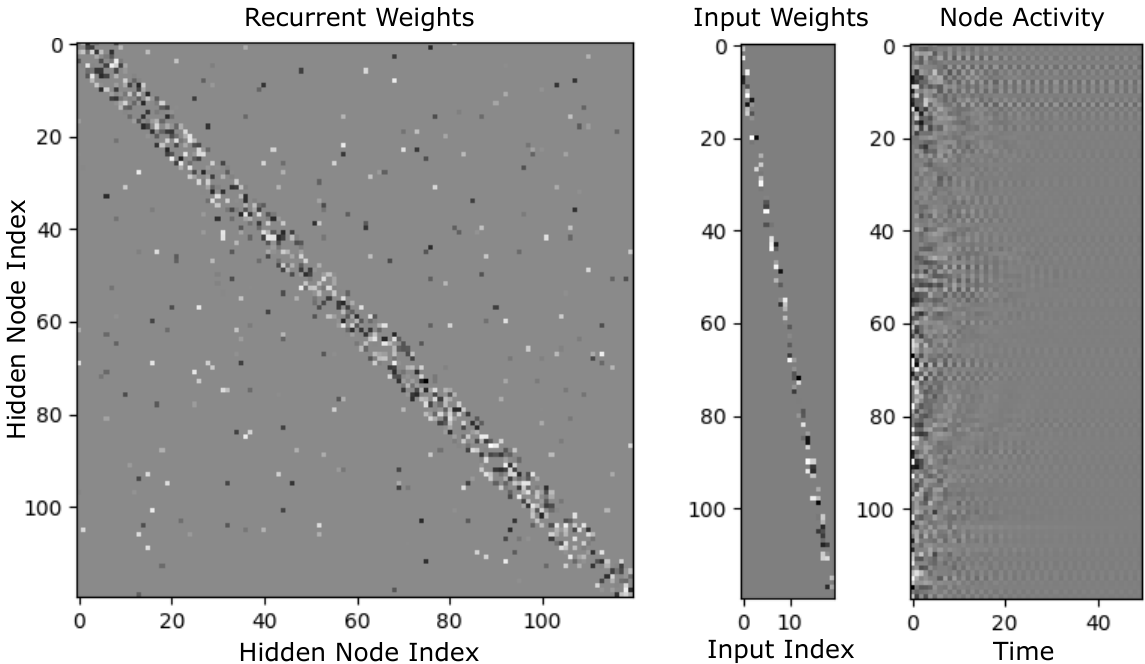}
    \caption{Left: The reservoir's recurrent weight matrix after sparsification to local connectivity. Center: sparsified input weight matrix such that each input sends to a fixed, dedicated number of unique and shared hidden nodes. Right: Resulting activity pattern of the reservoir after an initial random input vector followed by zero input for 50 steps.}
    \label{fig:weights}
\end{figure}

All agents followed the same model design shown in Figure \ref{fig:modeldiagram}.
Environmental data and agent feedbacks are passed as input to the recurrent memory module, and the recurrent state is passed to an actor and a critic module.
The feedback vector consists of the previous action, previous reward.
Previous action was coded as a one-hot vector.
Previous reward was discretized to its sign only, i.e. $\{-1, 0, 1\}$.
The critic module outputs only a scalar value of the current state.
The agent module outputs a categorical distribution from which the next action is sampled.
The hyperparameters for each model are given in Appendix A.
All models used the same actor-critic\cite{sutton2018reinforcement} training algorithm regularized by entropy maximization.

To get fair comparisons of memory architectures, results were obtained after specifying the recurrent portion of the network to be each of the following: 
(1) A linear layer, making the agent a simple multi-layer perceptron (MLP) as the null hypothesis,
(2) simple RNN,
(3) GRU,
(4) LSTM,
(5) ESN,
(6) ESN with dense local connectivity and sparse global connectivity.

\paragraph{Locally connected ESN}
The weight matrices of the locally connected ESN are shown in Figure \ref{fig:weights} 
The reservoir was designed by randomizing and sparsifying an RNN layer with the $\tanh$ activation function according to principles explained in {Luko{\v{s}}evi{\v{c}}ius and Jaeger. \cite{lukovsevivcius2009reservoir}
The reservoir layer is instantiated in terms of a fixed number of unique nodes dedicated to each input node ($N_\text{Unique}$) and a number of nodes shared per consecutive pair of inputs ($N_\text{Shared}$).
Input and recurrent weights are sparsified according to local and global probabilities.
For more details, see Appendix B: Connectivity Specifications.


\section*{Results}

\begin{figure}[ht]
    \centering
    \begin{subfigure}[t]{0.48\textwidth}
        \includegraphics[width=\linewidth]{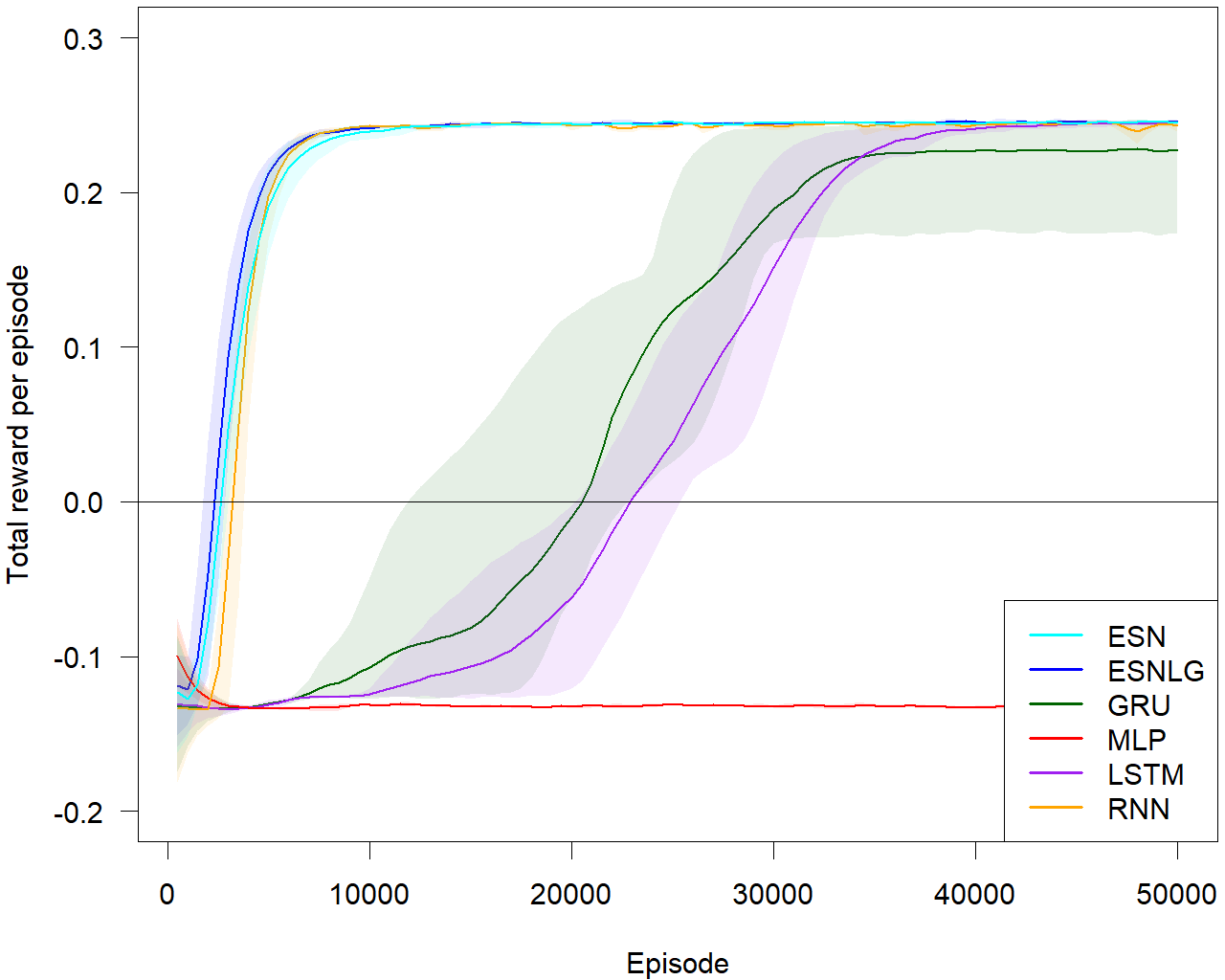}
        \caption{Results for recall match}
        \label{fig:results_rm}
    \end{subfigure}
    \begin{subfigure}[t]{0.48\textwidth}
        \includegraphics[width=\linewidth]{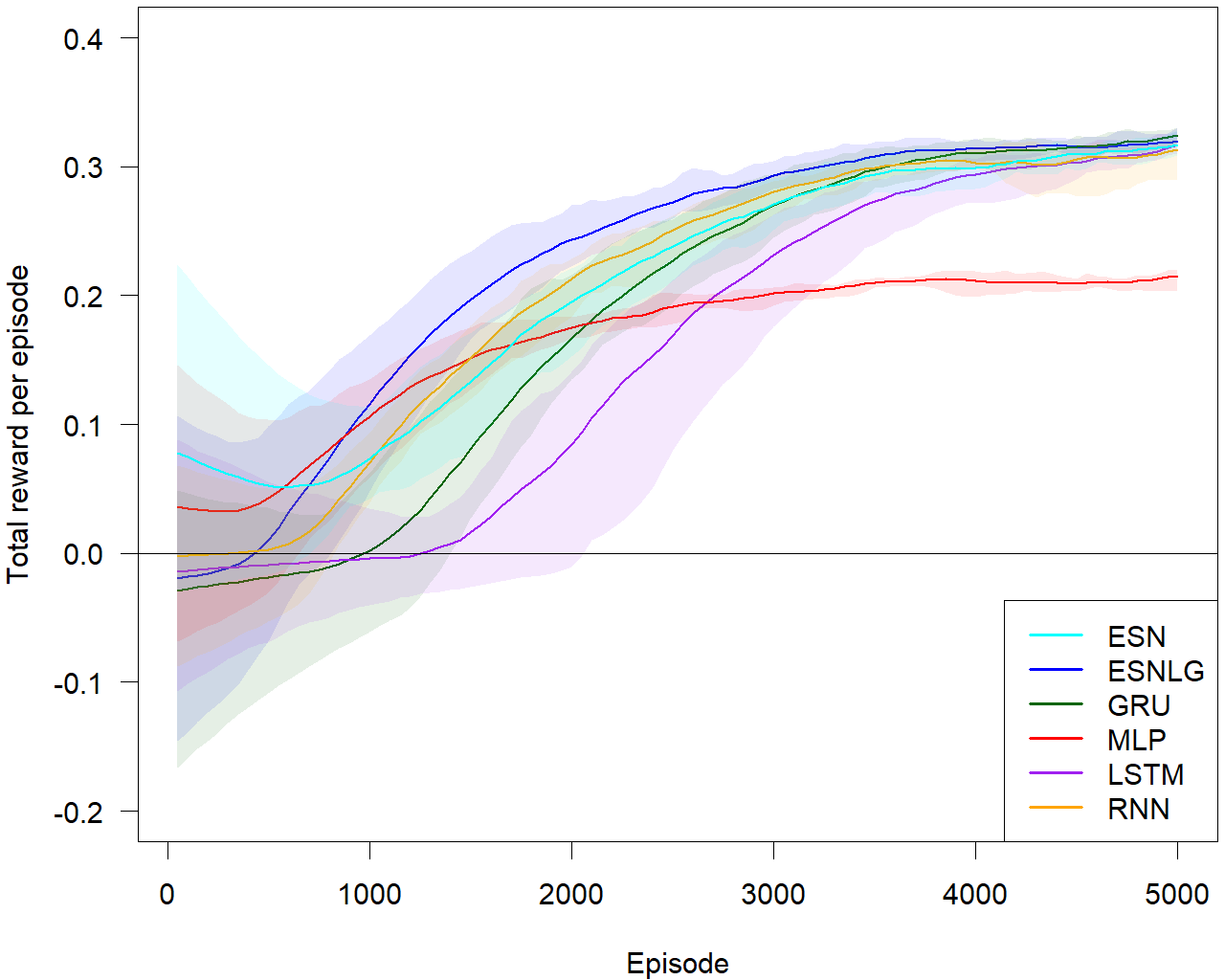}
        \caption{Results for multi-armed bandit}
        \label{fig:results_mab}
    \end{subfigure}

    \begin{subfigure}[t]{0.48\textwidth}
        \includegraphics[width=\linewidth]{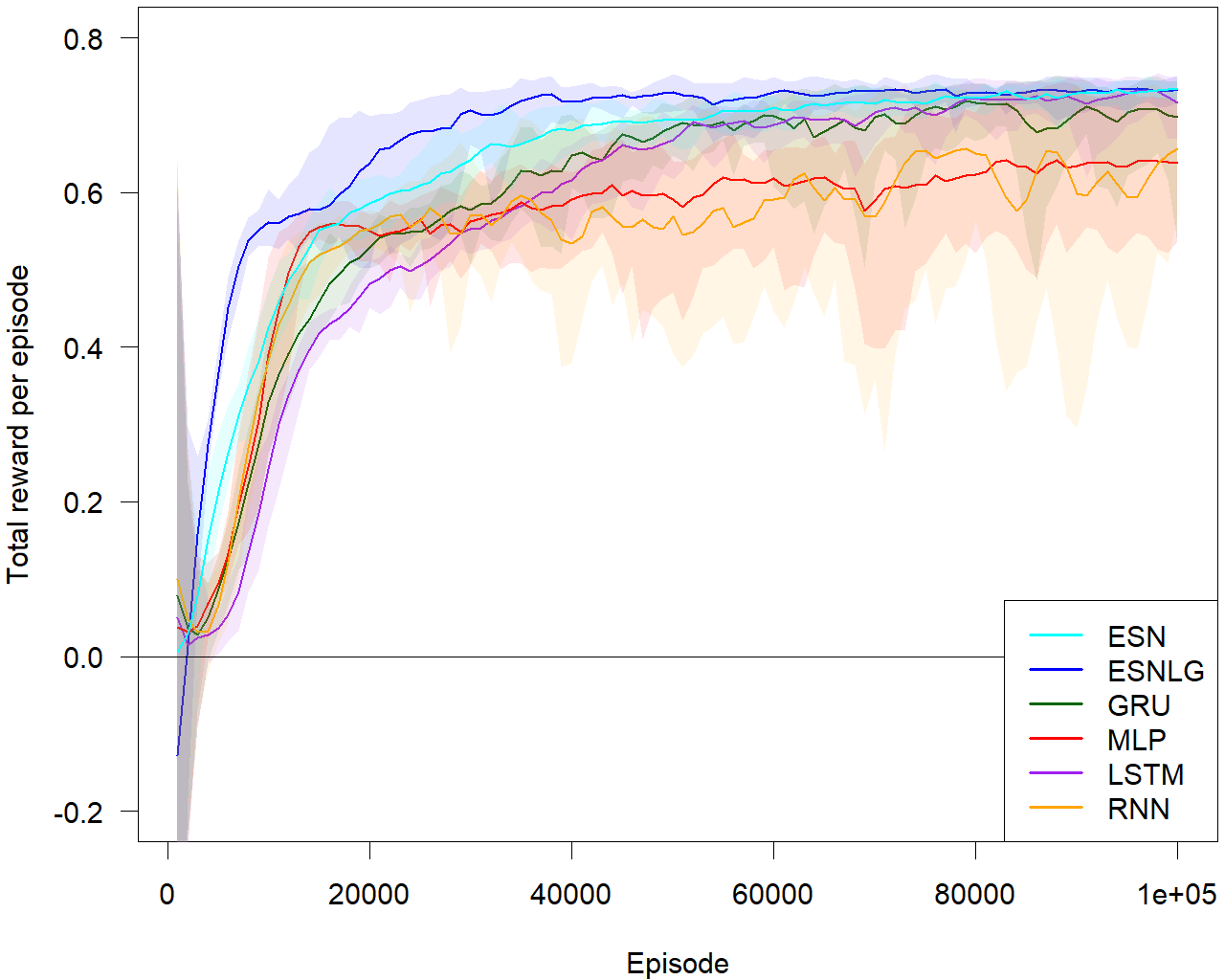}
        \caption{Results for water maze}
        \label{fig:results_wm}
    \end{subfigure}
    \begin{subfigure}[t]{0.48\textwidth}
        \includegraphics[width=\linewidth]{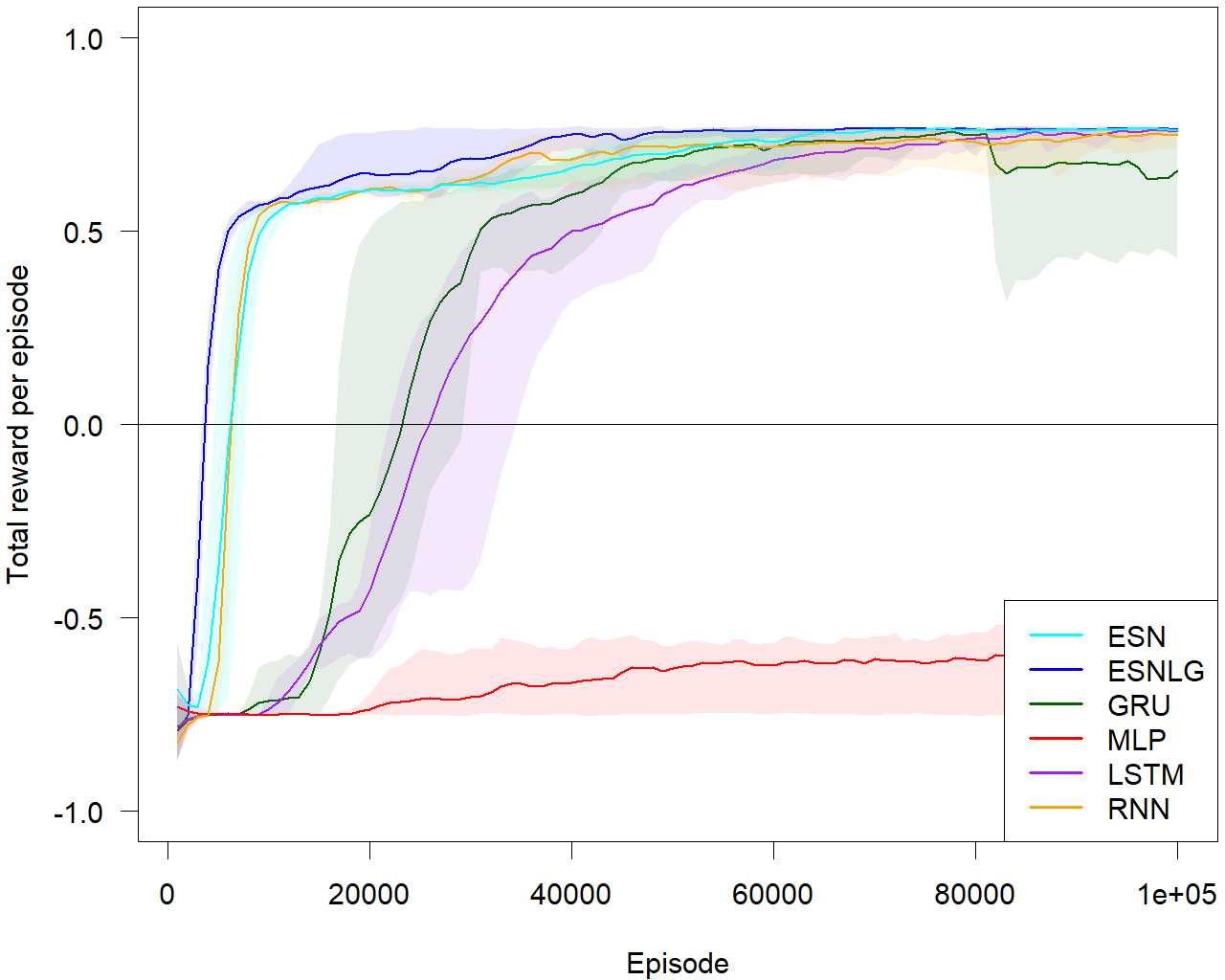}
        \caption{Results for Sequential Bandit}
        \label{fig:results_sb}
    \end{subfigure}
    \caption{Model comparisons on four memory POMDP tasks. Each model was run 8 times per task. Colored intervals show the minimum to maximum reward per time step. Lines show mean values.}
    \label{fig:results}
\end{figure}

\paragraph{Recall Match}
On recall match, despite the simplicity of the task, training required ten times as many episodes (50,000) for gated memory types than for the ESNs and the RNN.
This likely owes to the problem that gated units must learn what information to retain, which in this case is a large search space that is not benefitting from any inductive biases.
By comparison, both ESN types were fully trained in only about 5000 episodes because they required only learning to map the co-present traces of each observed symbol to the output.
As can be expected from this task design, the MLP was unable to learn the task, having no representation of observations two and four steps in the past. 
The ESNLG trained the fastest.

\paragraph{Multi-armed bandit}
All models learned the multi-armed bandit at different rates. 
The MLP learned to obtain reward only about two-thirds the rate of those with memory, as it had no source for learning a cumulative representation of reward probability, instead mapping only from the previous action.
The LSTM and GRU trained slower overall than the ESNs and RNN, with the ESNLG once again having the fastest training time.

\paragraph{Morris Water Maze}
Here too we see that the ESNLG learns to first locate the hidden reward and then exploit its memory of such significantly faster than the other methods.
The basic ESN and the gated memory units were able to achieve the same high score as the ESNLG but required about twice as many episodes to converge.
The MLP and the RNN were roughly matched, suggesting that both were able to learn efficient explore heuristics but unable to leverage long enough memory to re-locate the found reward in subsequent trials.
The RNN showed the greatest variability in performance near convergence. while the ESNLG showed the least variability.

\paragraph{Sequential bandit}
For sequential bandit, the same ranking of results occurred as for the other tasks. 
Like with recall match, there was a large disparity between the simpler architectures and the gated memory units.
The ESNs and RNN reached a local optimum within 10,000 episodes, then reached the global optimimum around 40,000.
The gated units gradually improved starting at 10,000 episodes and converging to the global optimum around 80,000 episodes with much greater variability than the former models.

\section*{Discussion}
The results of these experiments demonstrate that ESNs can not only produce major performance gains over trainable memory units, but that depending on how memory is required for the task, it can be extremely inefficient to train gated memory by reward signals.
This is most apparent in the results on Recall Match, which despite its simplicity, required an order of magnitude more training for the gated units than the ESNs or the RNN.

Second to the ESNs, we find that the RNN was also an efficient choice on these tasks but reached the solutions more slowly than the ESNs.
Because of this, we cannot provide any justification for training RNNs.
Our results may be explained by the fact that the RNN's weights were not initialized to be sparse or to have a spectral radius of 1.0, and so training likely consisted in just achieving some similar configuration from the imperfect starting point of random, normally distributed weights.

In each case, the GRU outperformed the LSTM in training efficiency.
In fact, the results perfectly rank by the number of free parameters in the recurrent module, supporting the intuition that free parameters controlling recurrent processing are a major source of inefficiency and complexity.
As the hyperparameters were chosen per model to all have roughly the same number of parameters, the performance disparities cannot be attributed to overall differences in the dimensionality of the solutions.

A central motivation for this study was the concept of metalearning, as defined and demonstrated by Wang et al. \cite{wang2018prefrontal}
In this framing, there are no special mechanisms for learning to learn.
It is rather a basic outcome of training agents with memory on distributional tasks that can only be solved with some degree of exploration.
In this view, a large-scale metalearner needs only the best possible memory systems in terms of efficiency and scale.
From these results, it appears that ESNs are an excellent choice for short-term memory system for rapid, expansive, and large-scale metalearning systems.
Successive work will attempt to incorporate methods of arbitrarily long-term memory with ESNs.

\paragraph{Limitations}
Here we only tested ESNs, though many reservoir types are possible.
The most common recommendation from the literature is that for computing with conventional hardware and software (such as Pytorch \cite{paszke2019pytorch}), ESNs are the fastest and most effective choice.
In the moderate-activation regime, ESNs may behave similarly to linear reservoirs and hence not produce a wide variety of \textit{a priori} nonlinear computations, so the possibility for demonstrating such remains open.

This study only considered a few reasonable values of hyperparameters and did not seek to undertake extensive search.
The chosen hyperparameters for both reservoirs and network architecture have been justified elsewhere.
These results also do not specifically demonstrate robustness to different hyperparameters, though it is known that gradients backpropagated extensively over time are more sensitive to reward scale and learning rate than those backpropagated over only a few steps.

Another question this study does not address is the comparative benefits of each network with respect to massive scale, such as is required for language modeling or image and video generation.
The ESNs required many more nodes than the trainable recurrent modules, yet they also freed up many parameters which were then used to expand the dimensions of the decoder networks.
Compute times were similar and small across all models, but only in the case of reservoir computers is there the immediate possibility for an entirely different choice of hardware. \cite{tanaka2019recent, fernando2003pattern}

\subsection*{Conclusion}
This study aimed to consider whether ESNs present any unique benefits for reinforcement learning on tasks requiring memory. 
In some cases, the results show overwhelming advantages in training efficiency over gated units.
Simple RNNs were comparable but never better than reservoirs, suggesting that the added trainability offers nothing above and beyond the results of a good spectral radius and connectivity structure.
Localized connectivity was found in all instances to result in even faster training with the ESN, and indeed that model performed best in every comparison.

\bibliographystyle{unsrtnat}
\bibliography{refs}

\section{Appendix A: Hyperparameters}
The hyperparameters for the ESNs are given in Table \ref{tab:hyperpars}.

\begin{table}[!t]
    \centering
    \caption{Hyperparameters for all models, including fixed ESN weight matrices.}
    \begin{tabular}{l|l|l}
        \hline
        Parameter & Description & Value\\ 
        \hline
        \textbf{Actor-Critic} &  &\\
         $LR$ & Learning rate  & 0.0003\\
         $\beta_e$ & Entropy regularization coefficient & 0.001\\
         $N_\text{Hidden}$ & Hidden units per MLP layer & 32-62 \\
         $K$ & Number of hidden layers per MLP & 2 \\
         \hline
        \textbf{ESN} & &\\
         $N_\text{Hidden}$ & 64\\ 
         $\phi$ & Spectral radius & 1.0 \\
         $P_G(W)$ & Global connection probability & 40\% \\
         $P_I(W)$ & Input connection probability & 40\% \\
         \hline
          \textbf{ESN with Local Connectivity} & &\\
         $N_{\text Unique}$ & Unique hidden nodes per input & 20\\
         $N_{\text Shared}$ & Overlapping nodes per neighboring inputs & 10\\ 
         $\phi$ & Spectral radius & 1.0 \\
         $P_L(W)$ & Local connection probability & 50\% \\
         $P_G(W)$ & Global connection probability & 1\% \\
         $P_I(W)$ & Input connection probability & 50\% \\
         $R$ & Max local connection radius & 10 \\
         \hline
    \end{tabular}
    \label{tab:hyperpars}
\end{table}

\section{Appendix B: Connectivity specifications}
Several operations are performed to prepare the fixed weight matrices of the locally connected ESN.
First, the recurrent weight matrix $W$ is masked to only include the diagonal and the first $R$ upper and lower off-diagonals.
This produces a local connectivity structure, treating position in the vector as a spatial location.
Second, the remaining non-zero weights are further masked with a random binary matrix, where elements equalled one with probability $P_L(W)$.
Third, global weights are added back in by first generating a new random weight matrix, and performing the same random binary masking with probability $P_G(W)$. 
The remaining non-zero elements of this global weight matrix replace zeroed elements of $W$ in the same respective indices.
Fourth, the spectral radius of $W$ is computed by taking the absolute value of its largest eigenvalue. 
The entire matrix is then element-wise divided by this value, then multiplied by hyperparameter $\phi$ to assign that as the new spectral radius.
In this way, the dynamical behaviors of the reservoir that control its memory properties can be deliberately specified.
Last, the input weight matrix is also masked such that each input projects only to $N_\text{Unique} + N_\text{Shared}$ nodes.
It is then sparsified with a random binary mask of probability $P_I(W)$.
The shared nodes are such because they take inputs from inputs with neighboring indices.
All bias terms in the reservoir were set to zero.

\end{document}